\pdfoutput=1
\documentclass[11pt]{article}

\usepackage{emnlp2021}

\usepackage{times}
\usepackage{latexsym}

\usepackage[T1]{fontenc}
\usepackage[utf8]{inputenc}

\usepackage{microtype}

\usepackage{hyperref}
\usepackage{graphicx}	% For figure environment
\usepackage[flushleft]{threeparttable}
\usepackage{float}
\usepackage{tcolorbox}

\title{Legal Transformer Models May Not Always Help}

\author{Saibo Geng \\
  EPFL,  Switzerland \\
  \texttt{saibo.geng@epfl.ch} \\\And
  Rémi Lebret \\
  EPFL, Switzerland \\
  \texttt{remi.lebret@epfl.ch} \\\And
  Karl Aberer \\
  EPFL, Switzerland \\
  \texttt{karl.aberer@epfl.ch} \\}

\begin{document}

\maketitle

\begin{abstract}
  Deep learning-based Natural Language Processing methods, especially transformers, have achieved impressive performance in the last few years. 
  Applying those state-of-the-art NLP methods to legal activities to automate or simplify some simple work is of great value.  
  This work investigates the value of domain adaptive pre-training and language adapters in legal NLP tasks.  
  By comparing the performance of language models with domain adaptive pre-training on different tasks and different dataset splits, we show that domain adaptive pre-training is only helpful with low-resource downstream tasks, thus far from being a panacea.
  We also benchmark the performance of adapters in a typical legal NLP task and show that they can yield similar performance to full model tuning with much smaller training costs. 
  As an additional result, we release LegalRoBERTa, a RoBERTa model further pre-trained on legal corpora.
\end{abstract}

\section{Introduction}

The adoption of natural language processing in the legal domain has a long history. The earliest systems for searching online legal content appeared in the 1960s and 1970s, and legal expert systems were a hot topic of discussion in the 1970s and 1980s.\cite{dale_2019} NLP has been applied to various legal areas to automate activities, including Natural Language Understanding based Contract review, question-answering based Legal advice \cite{dale_2019}. Various Natural Language tasks can be adapted to the legal domain, including

\begin{itemize}

\item Question Answering 
\item Argument detection, Definition Extraction
\item Semantic Annotation
\item Classification

\end{itemize}

In 2018, the release of BERT \cite{devlin2019bert} as a pre-trained language representation has achieved new state-of-the-art results on various NLP tasks. 
Later on, domain-adaptive pretraining(DAPT) and task-adaptive pretraining(TAPT)\cite{gururangan2020dont} on the pre-trained model can further improve the results.
Based on this idea, researchers have attempted to conduct DAPT with legal corpora, LegalBERT from \cite{chalkidis2020legalbert} is a good example. 
Later,\cite{liu2019roberta} argued that BERT is largely under-trained and suggested a new setup of key hyper-parameters and training schema to produce RoBERTa.
RoBERTa significantly outperforms BERT's performance on various tasks.
Following the same idea as LegalBERT, we believe a further domain adaptation with RoBERTa should produce a new state-of-the-art model in the legal domain.
We present LegalRoBERTa, a RoBERTa model further pre-trained on legal corpora.

However, unsupervised pretraining is a costly action that may take more than several days, even with good computing resources.
\cite{chalkidis2020legalbert} showed that compared to BERT, legalBERT achieves only slightly better performances on three different legal NLP tasks:

\begin{itemize}
  \item Multilabel Classification task on EURLEX57K dataset \cite{chalkidis-etal-2019-large} 
  \item Multilabel Classification task on ECHR-CASES dataset \cite{chalkidis-etal-2019-large} 
  \item Named Entity Recognition on CONTRACTS-NER dataset \cite{chalkidis2021neural}
\end{itemize}

\cite{compact_langauge_model} reported that corpus-specific MLM was not beneficial on a French QuAD task.
If the improvement from DAPT is not apparent compared to the original model, it is doubtful whether researchers should spend time collecting data and conducting DAPT every time a new language model is available.
We tested the popular open-source legal language models, including LegalRoBERTa produced by ourselves, to see if a significant improvement could be observed.
As we will show in the section \ref{sec:Downstream Legal NLP Tasks and Model Testing}, both legalBERT and legalRoBERTa demonstrated a limited boost compared to the original language model in the legal text classification task.

\cite{Pretrain_or_Not} pointed out that the improvement brought by the pre-training was related to the data size of downstream tasks.
The improvement is more remarkable when the downstream task is low-resource and vice-versa. 
As DAPT is also a particular type of pre-training; we believe that the benefit of DAPT may also have similar relationships with downstream tasks.
To demonstrate this hypothesis, we tested legal language models on two different tasks: one rich resource, one low resource as well as the same task with different sizes of training data.
The results showed that our hypothesis held and DAPT was especially beneficial while downstream task suffers from a lack of data.

Finally, we investigate the performance of a new NLP technique \textit{adapters} in legal NLP tasks.
The adapter is a more efficient way to fine-tune pre-trained language models to downstream tasks. 
It is faster to train and takes less space on disk. 
Our experimental results showed that the adapter was able to produce a comparable performance as fine-tuning the full model.
\section{Contributions}

The contributions of this paper are: 
\begin{enumerate}
  \item 
  Inspired by the idea of legalBERT and the success of RoBERTa, we present legalRoBERTa, a domain-adapted language representation for the legal area. 
  It was pre-trained on less legal corpora than legalBERT but produced a similar performance as legalBERT. 
  \item We adapted an existing legal summarization dataset \cite{aus} to a legal text retrieval task.
  \item We demonstrated that current open-source legal language models could only bring marginal benefit or no improvement on a rich resource NLP task. On the other hand, We showed that DAPT was beneficial when the downstream task was low resource. 
  \item We tested the performance of adapters \cite{houlsby2019parameterefficient} on a legal text classification task and show that they can produce comparable results as fine-tuning the full model.

\end{enumerate}

\section{Related Work}
\cite{chalkidis2020legalbert} introduced legalBERT as the first transformer adapted to the legal domain. 
Our work is trying to fill the gap between BERT and RoBERTa in the legal area and investigate whether DAPT is truly helpful in legal NLP tasks.
\cite{When_Does_Pretraining_Help} reported a similar conclusion on the relation of DAPT and downstream tasks. Our work has been done simultaneously with them, and we were not aware of their results until the work is mostly finished.

\section{Language Model Pre-training: LegalRoBERTa}
\subsection{Legal Corpora Description}

\begin{table}[!ht]
    \centering
    \small
    \begin{tabular}{c|c|c}
    \hline
         Corpus &  Size(raw) & Size(clean)   \\
    \hline
         Patent Litigations & 1.57GB & 1.1GB  \\
    \hline
         Caselaw Access Project & 5.6GB & 2.8GB \\
    \hline
         Google Patents Public Data &1.1GB & 1.0GB  \\
    \hline
         Total & 8.3GB & 4.9GB  \\
    \hline
    \end{tabular}
    \caption{Pre-trained corpora }
    \label{tab:sub_corpus_size}
\end{table}

As the first step to build a legal language model, we tried to collect public law-related corpora, but there were minimal available resources. 
As legal documents could contain sensitive information, institutes usually only release a small part of the data to the public, and a portion of them are in PDF format. 
Finally, we obtained around 5 GB of clean legal text data to proceed with the domain-adaptive pre-training.

\subsection{Comparison with Other corpora}

\begin{table}[!ht]
    \centering
    \small
    \begin{tabular}{c|c|c|c}
    \hline
         Domain&  Corpus & \#Token & Size(GB)   \\
    \hline
         BIOMED &S2ORC & 7.55B &47   \\
    \hline
         CS &S2ORC & 8.10B &48   \\
    \hline
         NEWS & REALNEWS  & 6.66B &39   \\
    \hline
         REVIEW & AMAZON reviews & 2.11B & 11   \\
    \hline
         LEGAL & LEGAL-BERT & - & 12   \\
    \hline
         \textbf{LEGAL} & \textbf{LEGAL-ROBERTA} & \textbf{1.01B} & \textbf{4.9}   \\
    \hline
    \end{tabular}
    \caption{Corpora for various domain }
    \label{tab:various_domain_corpora_size}
\end{table}

Compared to other domain adaptive pre-training experiments, our legal corpora is significantly smaller.

\subsection{Pre-training Details}

Following \cite{devlin2019bert}, we run additional pre-training steps of RoBERTa-BASE on domain-specific corpora. 
While \cite{devlin2019bert} suggested additional steps up to 100k, our pre-training goes up to 446k as \cite{chalkidis2020legalbert} suggests that prolonged in-domain pre-training brings a positive effect to future fine-tuning on downstream tasks. 

Fine-tuning configuration:
\begin{itemize}
    \item learning rate = 5e-5(with learning rate decay, ends at 4.95e-8)
    \item number of epochs = 3
    \item Total steps = 446K
    \item Total flops = 2.7365e18
    \item Device: 2*GeForce GTX TITAN X (computeCapability= 5.2 )
    \item RunTime 101 hours
    \item Per GPU batch size = 2
\end{itemize}

Loss starts at 1.850 and ends at 0.880
The perplexity on legal corpus after domain adaptive pre-training = 2.2735

However, given limited graphical memory space, our batch size is significantly smaller than those used in similar domain adaptive pre-training. We think further pre-training on the legal corpora should be considered and may be beneficial, c.f. Table \ref{app:pre-training details}. 
RoBERTa-BASE has been pre-trained for significantly more steps(1M) in generic corpora (e.g., Wikipedia, Children’s Books); thus, it is highly skewed towards generic language.\cite{chalkidis2020legalbert}
In Appendix \ref{app:next-token-prediction}, we give two concrete examples to demonstrate how differently is LegalRoBERTa behaving against RoBERTa and LegalBERT from \cite{chalkidis2020legalbert}  performance on Next-Token-Prediction task.

\section{Downstream Legal NLP Tasks and Model Testing}
\label{sec:Downstream Legal NLP Tasks and Model Testing}

\begin{table}[!ht]
  \centering
  \resizebox{\columnwidth}{!}{%
  \begin{tabular}{lll}
  \hline
    Model & Authors &  HuggingFace url  \\
  \hline
    Legal-BERT & \cite{chalkidis2020legalbert} & nlpaueb/legal-bert-base-uncased   \\
  \hline
    LegalRoBERTa & Our paper & saibo/legal-roberta-base \\
  \hline
    LegalBERT & \cite{When_Does_Pretraining_Help} & zlucia/legalbert \\
  \hline
    RoBERTa-base & \cite{liu2019roberta} &  roberta-base \\
  \hline
    BERT-base-unc & \cite{devlin2019bert} & bert-base-uncased  \\
  \hline
    rand-RoBERTa & Our paper & saibo/random-roberta-base \\
  \hline
  \end{tabular}%
}
  \caption{Tested Language Models with HuggingFace URLs}
  \label{tab:Tested Language Models}
\end{table}

\begin{table*}[!ht]
  \centering
\begin{threeparttable}
  \begin{tabular}{c|c|c|c|c|c|c|c}
  \hline
    Model & Precision &  Recall & F1 & R@5 & P@5 & RP@5 & NDCG@5  \\
  \hline
    Legal-BERT & \textbf{0.86} & 0.63 & 0.73 & \textbf{0.72} &0.69 & \textbf{0.79} & \textbf{0.82}  \\
  \hline
    LegalRoBERTa &0.84 & 0.63 & 0.72  & 0.70 &  0.67 & 0.78 & \textbf{0.80} \\
  \hline
    LegalBERT & \textbf{0.86} & 0.61 & 0.71  & 0.71 & 0.68 &0.78 & 0.81 \\
  \hline
    RoBERTa-base & 0.85 & \textbf{0.65} & \textbf{0.74}  & \textbf{0.72} & 0.69 & \textbf{0.79} & \textbf{0.82} \\
  \hline
    BERT-base-uncased & \textbf{0.86} & 0.62 & 0.72  & 0.72 & 0.69 &0.79 & 0.82 \\
  \hline
    random-RoBERTa & \textbf{0.85} & 0.59 & 0.69  & 0.69 & 0.66 &0.76 & 0.79 \\
  \hline
  \end{tabular}
  \caption{Results on Large-Scale Multi-Label Text Classification on EU Legislation} 
  \begin{tablenotes}
    \item Only Legal-BERT from \cite{chalkidis2020legalbert} has slightly outperformed original BERT. 
    The adapted model from \cite{When_Does_Pretraining_Help} is slightly below the original BERT, 
    so is legalRoBERTa against original RoBERTa.
\end{tablenotes}
  \label{tab:lmtc experimental results}
\end{threeparttable}
\end{table*}

To investigate whether legal language models are better compared with normal language models. We selected  six open-source language models available on HuggingFace, including two adapted BERT models from different authors: Legal-BERT from \cite{chalkidis2020legalbert} and LegalBERT from \cite{When_Does_Pretraining_Help}, one adapted RoBERTa model from us, original BERT and RoBERTa models, and finally, a randomly initialized RoBERTa model for comparison.  
All six models are available via HuggingFace API. Their links on HuggingFace are listed in the Table \ref{tab:Tested Language Models}.

We evaluated these models on text classification and information retrieval using two different datasets.
EURLEX57K \cite{chalkidis-etal-2019-large} is a large-scale multi-label text classification(LMTC) dataset of EU laws.
Legal Case Reports Data Set \cite{Dua:2019} is a dataset containing Australian legal cases from the Federal Court of Australia (FCA) during 2006-2009, which was built to experiment with automatic summarization and citation analysis.

\begin{table}[!ht]
  \centering
  \small
  \begin{tabular}{c|c|c|c}
  \hline
      Split & Documents(D) &  Words/D &  Labels/D   \\
  \hline
      Train & 45k & 729  & 5    \\
  \hline
      Dev & 6k & 714 & 5   \\
  \hline
      Test & 6k & 725 & 5  \\
  \hline
      Total & 57k & 727 & 5  \\
  \hline
  \end{tabular}
  \caption{Statistics of the EUR-LEX dataset}
  \label{tab:Statistics of the EUR-LEX dataset}
\end{table}

\subsection{Large-Scale Multi-Label Text Classification on EU Legislation(rich-resource)}

\textbf{Experimental setup}\par
Our dataset in this task contains 57K legislative documents from EUR-LEX, annotated with around 4.3K labels. 
Each document can be labeled to more than one label; thus, it is a multi-label classification task.
The 4,271 labels are divided into frequent (746 labels), few-shot (3,362), and zero-shot (163), depending on whether they were assigned to more than 50, fewer than 50 but at least one, or no training documents, respectively. 
The model is composed of a language model as encoder and an extra classification layer on top. 
We use binary cross-entropy as the loss function in this task.The metrics we used in this task are identical as in the paper \cite{chalkidis-etal-2019-large}.

\begin{table*}[!ht]
  \centering
  \small
  \begin{tabular}{c|c|c|c|c|c|c|c|c|c}
  \hline
    Train data ratio & Train samples & Model & Precision &  Recall & F1 & R@5 & P@5 & RP@5 & NDCG@5  \\
  \hline
    100\% & 45000& BERT& 0.86 & 0.62 & 0.72 & 0.72 &0.69 & 0.79 & 0.82  \\
  
          &      & DAPT improvement & +0.00 & +0.00 & +0.01 & +0.00 & +0.00 & +0.00 & +0.00  \\
          &      & relative improvement(\%)          & 0.0 & 0.0 & 1.4 & 0.0 & 0.0 & 0.0 & 0.0  \\
  \hline
    20\% & 9000 &BERT       &0.66 & 0.19 & 0.29  & 0.39 &  0.35 & 0.43 & 0.46 \\
         &      & DAPT improvement & +0.04 & +0.00 & +0.00 & 0.01  &  0.00 & 0.00 & +0.00 \\
         &      & relative improvement(\%)          & +6.1 & 0.0 & 0.0 & +2.8 & 0.0 & 0.0 & 0.0  \\
  \hline
    10\% & 4500 &BERT      &0.58 & 0.09 & 0.15  & 0.30 & 0.27 &0.33 & 0.35 \\
         &      & DAPT improvement & +0.06 & +0.02 & +0.03 & 0.02  &  0.01 & 0.01 & +0.02 \\
         &      & relative improvement(\%)           & +10.3 & \textbf{+22} & \textbf{+20} & +6.7 & +3.7 & +3.0 & +5.7  \\
  \hline
    5\% &  2250 &BERT       &0.49 & 0.06 & 0.11  & 0.22 & 0.20 & 0.24 & 0.26 \\
        &      & DAPT improvement & +0.0 & +0.01 & +0.02 & 0.02  &  0.02 & 0.02 & +0.02 \\
        &      & relative improvement(\%)           & 0.0 & \textbf{+17} & \textbf{+18} & +9.1 & \textbf{+10} & +8.3 & +7.7  \\
  \hline
    1\% & 450  &BERT       & 0.00 & 0.00  & 0.00 & 0.03 &0.02 & 0.03 & 0.02\\
        &      & DAPT improvement & +0.0 & +0.00 & +0.00 & 0.01  &  0.01 & 0.01 & +0.02 \\
        &      & relative improvement(\%)           & 0.0 & 0.0 & 0.0 & \textbf{+33} & \textbf{+50} & \textbf{+33} & \textbf{+100}  \\
  \hline
  \end{tabular}
  \caption{Improvement on LMTC with different sizes of training data}
  \label{tab:lmtc_size experimental improvement}
\end{table*}

\textbf{Results}\par
As shown in the Table \ref{tab:lmtc experimental results} only Legal-BERT from \cite{chalkidis2020legalbert} has slightly outperformed the original BERT. The adapted model LegalBERT from \cite{When_Does_Pretraining_Help} is slightly below the original BERT, so is LegalRoBERTa against original RoBERTa.
This slight margin could be due to statistical error because we only ran the experiments once with a fixed random seed. However, we could not pretend that pre-trained models have demonstrated significantly better performance than the general language models. 
\cite{chalkidis2020legalbert} reported similar results to us on the same task: the difference between legalBERT and BERT was \textbf{not significant}. 
One can also notice that the difference between pre-trained RoBERTa and randomly initialized RoBERTa is also relatively small. 
This is coherent with the argument from \cite{Pretrain_or_Not}: pretraining(or transfer learning in general) is only beneficial when there is not enough data for downstream tasks.

\subsection{Automatic Catchphrase Retrieval from Legal Court Case Documents(low-resource)} 
\label{sec:retrieval}

\begin{table}[!ht]
  \centering
  \small
  \begin{tabular}{c|c}
  \hline
      Split & Cases   \\
  \hline
      Train & 2807     \\
  \hline
      Dev & 350   \\
  \hline
      Test & 350   \\
  \hline
      Total & 3507   \\
  \hline
  \end{tabular}
  \caption{Statistics of the AUS-CASE dataset}
  \label{tab:Statistics of the AUS dataset}
\end{table}

\textbf{Experimental setup}\par

The dataset\cite{aus} contains 3886 Australian legal cases from the Federal Court of Australia (FCA). 
The cases are all annotated with catchphrases, citations sentences, citation catchphrases, and citation classes. 
The content which interests us most is the case sentences and catchphrases. 
Initially, the catchphrases were used as the gold standard for summarization experiments.
We decide to adopt this dataset to a text retrieval task because BERT as a bidirectional model is not well suited for text generation tasks, including summarization.
Our goal here is to retrieve the correct catchphrases based on the description of a case(sentences). 

We use a supervised approach to rank and retrieve catchphrases from court case documents. 
Our method is inspired by \textit{VSE++: Improving Visual-Semantic Embeddings with Hard Negatives} from  \cite{faghri2018vse}. 
In that work, image-caption retrieval was conducted using image-caption embedding similarity rank, on which the model is optimized via the minimization of triplet ranking loss. 
We applied the same idea to our catchphrase retrieval task by replacing the images with case descriptions and captions with catchphrases. 
An essential difference between this task and the previous task is the volume of training data.
Comparing the Table \ref{tab:Statistics of the AUS dataset} and Table \ref{tab:Statistics of the EUR-LEX dataset}, the retrieval task has only $1/10$ training data as the text classification task. 

As shown in the Figure \ref{app:retrieval_model}, the model in this task consists of an encoder (pre-trained transformer in our case) to extract features from both catchphrases and documents plus a dense neural network to transform features to representation in a shared space.  
The loss function in this task is the triplet loss function. 
Moreover, as we have only one ground true catchphrase, our task is roughly five times more difficult than the popular MS-COCO image-captioning task, where each image has five ground true captions on average.

%>>>legalroberta = pd.read_csv("./legalroberta.csv",index_col=0)
%>>>legalroberta.iloc[:,0:4].T.describe()
%

The metric we used in this task is Recall@K (R@K), Mean Rank of the ground-true label, and median rank of the ground-true label(MeanRank and MedRanks are the lower, the better). 
We notice that this metric also depends on the size of the test database. For example, searching over a one-thousand cases test set is more challenging than a one-hundred cases test set. 
In this project, we focus on the test set of size = 389 cases. 

\begin{table}[!ht]
  \centering
  \small
  % A table with adjusted row and column spacings
% \setlength sets the horizontal (column) spacing
% \arraystretch sets the vertical (row) spacing
% The \begingroup ... \endgroup pair ensures the separation
% parameters only affect this particular table, and not any
% sebsequent ones in the document.
  \begingroup 
  \setlength{\tabcolsep}{3pt} % Default value: 6pt
  \begin{tabular}{cccccc}
  \hline
      Model & R1 &  R5 & R10  & MedRank & MeanRank   \\
  \hline
      BERT & 14.4 & 33.7 & 45.7 & 13.0  & 49.8 \\
           & +-0.6 & +-0.8 & +-0.4 & +-0.0  & +-1.1 \\
  \hline
      RoBERTa & 13.6 & 33.1 & 44.7 & 14.0  & 55.5 \\
              & +-0.6 & +-0.8 & +-0.9 & +-0.0  & +-2.9 \\
  \hline
      Legal-BERT  & 16.8 & 37.7 & 49.9 & 10.6  & 46.9 \\
                     & +-1.0     &  +-0.5 & +-0.3 & +-0.6  & +-1.3 \\
  \hline
      LegalBERT  & 15.4 & 34.8 & 46.2 & 12.8  & 47.3 \\
                     &     +-0.2 &    +-1.3 &     +-1.1 &    +-0.4  &    +-1.8 \\
  \hline
      LegalRoBERTa  & 15.1 & 34.6 & 46.3 & 13.5 & 52.9 \\
                  & +-1.1 & +-0.9 & +-0.8 & +-1 & +-0.4 \\ 
  \hline
  \end{tabular}
  \endgroup
  \caption{Results on Automatic Catchphrase Retrieval from Legal Court Case Documents}
  \label{tab:retrieval_results}
\end{table}

\textbf{Results} \par 
As showed in the Table \ref{tab:retrieval_results}, all the domain-pre-trained models have outperformed the corresponding original model. 
Legal-BERT from \cite{chalkidis2020legalbert} is significantly better than the original BERT.
This result suggested that domain-adapted models should be \textbf{favored} in the case of low-resource tasks.
 
\begin{table*}[!ht]
  \centering
  \small
  \begin{tabular}{c|c|c|c|c|c|c|c}
  \hline
    Model & Precision &  Recall & F1 & R@5 & P@5 & RP@5 & NDCG@5  \\
  \hline
    Legal-BERT & 0.86 & 0.63 & 0.73 & 0.72 &0.69 & 0.79 & 0.82  \\
    \hline
    Adapter (diff) & +0.01 & -0.03 & -0.02 & -0.01 & -0.01 & 0.0 & 0.0  \\
  \hline
    LegalRoBERTa &0.84 & 0.63 & 0.72  & 0.70 &  0.67 & 0.78 & 0.80 \\
  \hline
  Adapter (diff) &+0.02 & -0.04 & -0.02  & 0.0 &  0.0 & 0.0 & +0.01 \\
  \hline
    LegalBERT & 0.86 & 0.61 & 0.71  & 0.71 & 0.68 &0.78 & 0.81 \\
  \hline
  Adapter (diff) & -0.01 & 0.0 & 0.0  & 0.0 & 0.0 & +0.01 & +0.01 \\
  \hline
    RoBERTa-base & 0.85 & 0.65 & 0.74  & 0.72 & 0.69 & 0.79 & 0.82 \\
  \hline
  Adapter (diff) & +0.01 & -0.06 & -0.04  & -0.01 & -0.01 & 0.00 & -0.01 \\
  \hline
  \end{tabular}
  \caption{Performance of adapters on LMTC task}
  \label{tab:adapter results}
\end{table*}

\section{Varying Downstream task data size}

Based on the observation from the previous experiments, we further investigate the relation between domain-pretraining and downstream task data size.
We vary the size of training data of LMTC tasks and compare the performance of BERT versus LegalBERT (see Table~\ref{app:lmtc}).

The results in Table \ref{tab:lmtc_size experimental improvement} show that domain-pretraining's performance improvement is tightly related to the downstream task training data size.
The benefit is significant when the training data is scarce and negligible when the training data is sufficient. 

\section{Adapters in legal text classification}

When there are multiple downstream tasks, it is time-wise inefficient to fine-tune the whole language model once per task, and it also requires much space to save the models for each task.
Adapter proposed by \cite{houlsby2019parameterefficient} is a good alternative to full model fine-tuning. 
Instead of updating all the parameters contained in the model, we add a so-called adapter module into the model.
In the training step, only parameters contained in the adapter module are updated while the language model itself remains unchanged. 
When training is finished, one only needs to save the adapter module for each task. 

This multi-task scenario is frequent in legal NLP tasks because a legal activity could be assisted by several relatively simple downstream tasks.
We test the adapter module on various language models with or without domain pre-training.
From the performance perspective, adapters can produce the same results as fine-tuning the whole model, cf Table \ref{tab:adapter results}. 
However, training adapters is not faster than fine-tuning the full model because the forward pass and back-propagation still have a similar amount of calculations as fine-tuning the whole model.

\section{Limitations and Future Work}

The size of legal corpora available restricted the pre-training of LegalRoBERTa. 
To better utilize the potential of RoBERTa, we should consider collecting more data such as automatic scraping. 
In the meantime, \cite{chalkidis2020legalbert} has released some other legal corpora of UK and EU legislative of roughly 2.5 GB. 
It should be beneficial to include those data into the pre-training of LegalRoBERTa v2. 
Furthermore, the pre-training steps seem to be insufficient compared with other related work (see~\ref{app:pre-training details}).
In the task of Large-Scale Multi-Label Text Classification on EU Legislation, we evaluated models with identical hyper-parameters due to limited time and limited computing resources. 
A grid search of hyper-parameters and repeated experiments several times with different random seeds could be considered.  

In the task of legal case retrieval, paired statistical testing can be conducted to conclude whether the domain pre-trained models are significantly better than the original models.

\section{Conclusion}
In this work, we first tried to answer a critical question in legal NLP from an empirical perspective: When does domain pre-training help the model to yield better performance? 
Through a series of legal NLP experiments, we showed that the existing three legal transformer models did not yield significant improvement on a rich-resource task while did show considerable 
improvement on a low-resource task or if we deliberately cut down the training data size. 
We therefore recommend domain pre-trained language models only in case of low-resource tasks.
The second part showed that adapters, as an emerging technique, are very suitable to solve legal NLP tasks.
As an intermediate result, we release LegalRoBERTa\footnote[1]{https://huggingface.co/saibo/legal-roberta-base}, a Roberta model adapted to the legal domain.

\section{Acknowledgments}
This work was done in cooperation with the Federal Department of Foreign Affairs(FDFA) of Switzerland and EPFL under a semester project.

\bibliographystyle{apalike}
\bibliography{main}

\begin{thebibliography}{}

\bibitem[Chalkidis et~al., 2019]{chalkidis-etal-2019-large}
Chalkidis, I., Fergadiotis, E., Malakasiotis, P., and Androutsopoulos, I.
  (2019).
\newblock Large-scale multi-label text classification on {EU} legislation.
\newblock In {\em Proceedings of the 57th Annual Meeting of the Association for
  Computational Linguistics}, pages 6314--6322, Florence, Italy. Association
  for Computational Linguistics.

\bibitem[Chalkidis et~al., 2020]{chalkidis2020legalbert}
Chalkidis, I., Fergadiotis, M., Malakasiotis, P., Aletras, N., and
  Androutsopoulos, I. (2020).
\newblock Legal-bert: The muppets straight out of law school.

\bibitem[Chalkidis et~al., 2021]{chalkidis2021neural}
Chalkidis, I., Fergadiotis, M., Malakasiotis, P., and Androutsopoulos, I.
  (2021).
\newblock Neural contract element extraction revisited: Letters from sesame
  street.

\bibitem[DALE, 2019]{dale_2019}
DALE, R. (2019).
\newblock Law and word order: Nlp in legal tech.
\newblock {\em Natural Language Engineering}, 25(1):211–217.

\bibitem[Devlin et~al., 2019]{devlin2019bert}
Devlin, J., Chang, M.-W., Lee, K., and Toutanova, K. (2019).
\newblock Bert: Pre-training of deep bidirectional transformers for language
  understanding.

\bibitem[Dua and Graff, 2017]{Dua:2019}
Dua, D. and Graff, C. (2017).
\newblock {UCI} machine learning repository.

\bibitem[Faghri et~al., 2018]{faghri2018vse}
Faghri, F., Fleet, D.~J., Kiros, J.~R., and Fidler, S. (2018).
\newblock Vse++: Improving visual-semantic embeddings with hard negatives.

\bibitem[Galgani and Hoffmann, 2010]{aus}
Galgani, F. and Hoffmann, A. (2010).
\newblock Lexa: Towards automatic legal citation classification.
\newblock In Li, J., editor, {\em AI 2010: Advances in Artificial
  Intelligence}, volume 6464 of {\em Lecture Notes in Computer Science}, pages
  445 --454. Springer Berlin Heidelberg.

\bibitem[Gururangan et~al., 2020]{gururangan2020dont}
Gururangan, S., Marasović, A., Swayamdipta, S., Lo, K., Beltagy, I., Downey,
  D., and Smith, N.~A. (2020).
\newblock Don't stop pretraining: Adapt language models to domains and tasks.

\bibitem[Houlsby et~al., 2019]{houlsby2019parameterefficient}
Houlsby, N., Giurgiu, A., Jastrzebski, S., Morrone, B., de~Laroussilhe, Q.,
  Gesmundo, A., Attariyan, M., and Gelly, S. (2019).
\newblock Parameter-efficient transfer learning for nlp.

\bibitem[Lee et~al., 2019]{biobert}
Lee, J., Yoon, W., Kim, S., Kim, D., Kim, S., So, C.~H., and Kang, J. (2019).
\newblock Biobert: a pre-trained biomedical language representation model for
  biomedical text mining.
\newblock {\em Bioinformatics}.

\bibitem[Liu et~al., 2019]{liu2019roberta}
Liu, Y., Ott, M., Goyal, N., Du, J., Joshi, M., Chen, D., Levy, O., Lewis, M.,
  Zettlemoyer, L., and Stoyanov, V. (2019).
\newblock Roberta: A robustly optimized bert pretraining approach.

\bibitem[Micheli et~al., 2020]{compact_langauge_model}
Micheli, V., d'Hoffschmidt, M., and Fleuret, F. (2020).
\newblock On the importance of pre-training data volume for compact language
  models.

\bibitem[Wang et~al., 2020]{Pretrain_or_Not}
Wang, S., Khabsa, M., and Ma, H. (2020).
\newblock To pretrain or not to pretrain: Examining the benefits of pretraining
  on resource rich tasks.
\newblock {\em CoRR}, abs/2006.08671.

\bibitem[Zheng et~al., 2021]{When_Does_Pretraining_Help}
Zheng, L., Guha, N., Anderson, B.~R., Henderson, P., and Ho, D.~E. (2021).
\newblock When does pretraining help? assessing self-supervised learning for
  law and the casehold dataset.
\newblock {\em CoRR}, abs/2104.08671.

\end{thebibliography}

\onecolumn

\section*{Appendix}

\subsection{Domain Adaptive Pre-training Details of Related Work}
\label{app:pre-training details}

In the table below, we can see the Domain Adaptive Pre-training details of related work.
\begin{table*}[!ht]
  \begin{tabular}{c|c|c|c|c}
  \hline
      Experiment & Authors &  Step & \# epoch & batch size  \\
  \hline
      Various Domains & \cite{gururangan2020dont} & 12.5K  & 1 & 256   \\
  \hline
      LegalBERT & \cite{chalkidis2020legalbert} & 1000K & 40 & 256 \\
  \hline
      BioBERT & \cite{biobert} & 200K-470K & - & 192 \\
  \hline
      LegalROBERTa & Our paper & 446K & 3 & 4 \\
  \hline
  \end{tabular}
\end{table*}

\subsection{ Examples of Next-Token-Prediction Results of LegalRoBERTa}
\label{app:next-token-prediction}

\begin{tcolorbox}
This \{\textbf{mask}\} Agreement is between General Motors and John Murray .
\end{tcolorbox}

\begin{table}[!h]
    \centering
    \small
    \begin{tabular}{|c|c|c|c|}
    \hline
        Model & Top1 &  Top2 & Top3  \\
    \hline
        BERT & new & current  & proposed\\
    \hline
        LegalBERT & settlement & letter & dealer \\
    \hline
        LegalROBERTa & License & Settlement & Contract\\
    \hline
    \end{tabular}
    \caption{Next Token Prediction Example 1}
    \label{tab:Next Token Prediction Example 1}
\end{table}

\begin{tcolorbox}
The applicant submitted that her husband was subjected to treatment amounting to \{\textbf{mask}\} whilst in the custody of Adana Security Directorate
\end{tcolorbox}

\begin{table}[!h]
    \centering
    \begin{tabular}{|c|c|c|c|}
    \hline
        Model & Top1 &  Top2 & Top3  \\
    \hline
        BERT & torture & rape  & abuse\\
    \hline
        LegalBERT & torture & detention & arrest \\
    \hline
        LegalROBERTa & torture & abuse & insanity\\
    \hline
    \end{tabular}
    \caption{Next Token Prediction Example 2}
    \label{tab:Next Token Prediction Example 2}
\end{table}

\subsection{Model Architecture in Catchphrase Retrieval task}
\label{app:retrieval_model}

Below is an illustration of the Automatic Catchphrase Retrieval from Legal Court Case Documents described in Section~\ref{sec:retrieval}.

\includegraphics[scale=0.40]{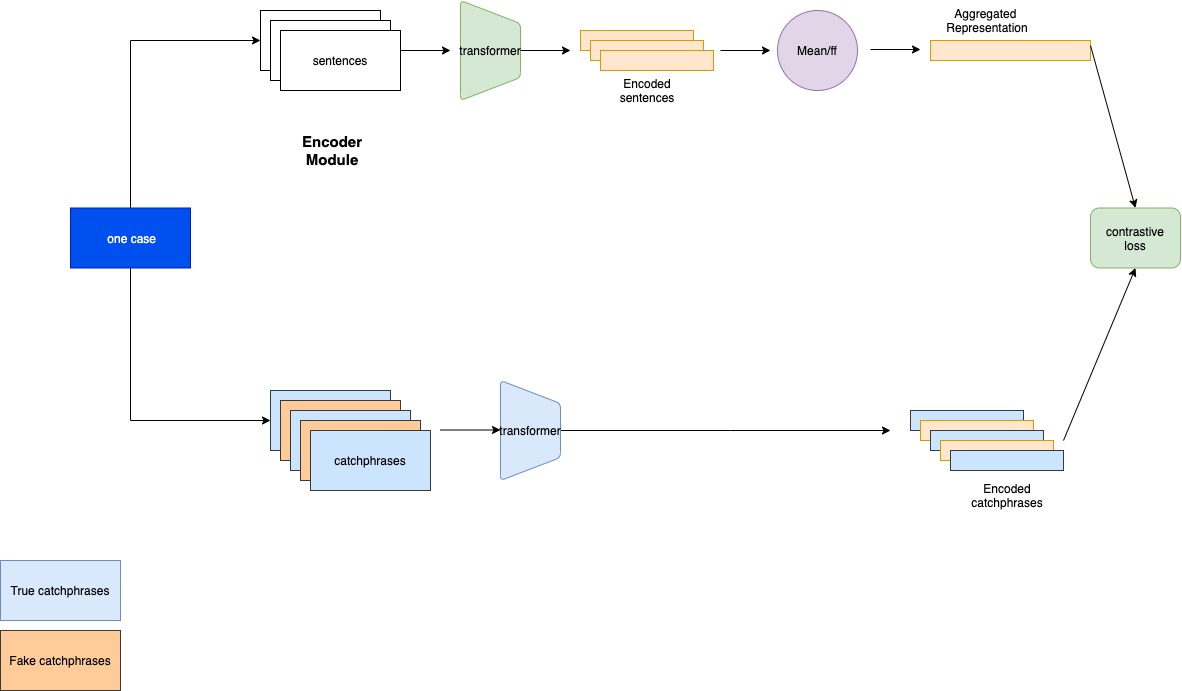}

\subsection{ Hyper-parameters in LMTC task}

\begin{enumerate}
  \item lr: 3e-05
  \item random seed: 0
  \item batch size: 16
  \item max sequence size: 216
  \item epochs: 40
  \item dropout:0.1
  \item early stop:yes
  \item patience: 7
\end{enumerate}

\subsection{Results on LMTC with different sizes of training data}
\label{app:lmtc}

The results on LMTC for BERT and LegalBERT with different sizes of training data are shown in the table below.

\begin{table*}[!ht]
\small
  \begin{tabular}{c|c|c|c|c|c|c|c|c|c}
  \hline
    Training data ratio & Train samples & Model & Precision &  Recall & F1 & R@5 & P@5 & RP@5 & NDCG@5  \\
  \hline
    100\% & 45000& BERT& 0.86 & 0.62 & 0.72 & 0.72 &0.69 & 0.79 & 0.82  \\
  
          &      & LegalBERT&0.86 & 0.63 & 0.73 & 0.72 &0.69 & 0.79 & 0.82  \\
  \hline
    20\% & 9000 &BERT       &0.66 & 0.19 & 0.29  & 0.39 &  0.35 & 0.43 & 0.46 \\
         &      & LegalBERT & 0.70 &0.19 & 0.29 & 0.40  &  0.35 & 0.43 & 0.46 \\
  \hline
    10\% & 4500 &BERT      &0.58 & 0.09 & 0.15  & 0.30 & 0.27 &0.33 & 0.35 \\
         &      & LegalBERT&0.64 & 0.11 & 0.18  & 0.32 & 0.28 &0.34 & 0.37 \\
  \hline
    5\% &  2250 &BERT       &0.49 & 0.06 & 0.11  & 0.22 & 0.20 & 0.24 & 0.26 \\
        &       & LegalBERT &0.49 & 0.07 & 0.13  & 0.24 & 0.22 & 0.26 & 0.28 \\
  \hline
    1\% & 450  &BERT       & 0.00 & 0.00  & 0.00 & 0.03 &0.02 & 0.03 & 0.02\\
        &      & LegalBERT & 0.00 & 0.00  & 0.00 & 0.04 &0.03 & 0.04 & 0.04\\
  \hline
  \end{tabular}
\end{table*}

\end{document}